\documentclass[10pt,conference]{IEEEtran}

\IEEEoverridecommandlockouts
\usepackage[T1]{fontenc}
\usepackage{cite}
\usepackage{amsmath,amssymb,amsfonts}
\usepackage{graphicx}
\usepackage{textcomp}
\usepackage{xcolor} 
\usepackage{subfigure}
\usepackage{siunitx}
\usepackage[noend]{algpseudocode}
\usepackage{algorithmicx,algorithm}
\usepackage{amssymb}
\usepackage{makecell}
\usepackage{balance}
\usepackage{color}

\usepackage[numbers,sort&compress]{natbib}
\makeatletter
\newcommand{\removelatexerror}{\let\@latex@error\@gobble}
\makeatother
 
\def\BibTeX{{\rm B\kern-.05em{\sc i\kern-.025em b}\kern-.08em
    T\kern-.1667em\lower.7ex\hbox{E}\kern-.125emX}}

\begin{document}

\title{Resource-Efficient Generative AI \\Model Deployment in Mobile Edge Networks}

\author{\IEEEauthorblockN{Yuxin~Liang\IEEEauthorrefmark{1},~Peng~Yang\IEEEauthorrefmark{1},~Yuanyuan~He\IEEEauthorrefmark{2}, and~Feng~Lyu\IEEEauthorrefmark{3}
\IEEEauthorblockA{\IEEEauthorrefmark{1}School of Electronic Information and Communications, Huazhong University of Science and Technology, Wuhan, China\\
\IEEEauthorrefmark{2}School of Cyber Science and Engineering, Huazhong University of Science and Technology, Wuhan, China\\
\IEEEauthorrefmark{3}School of Computer Science and Engineering, Central South University, Changsha, China\\
Email:\IEEEauthorrefmark{1}\{yuxinliang, yangpeng\}@hust.edu.cn, \IEEEauthorrefmark{2}yuanyuan\_cse@hust.edu.cn, \IEEEauthorrefmark{3}fenglyu@csu.edu.cn }}}

\maketitle
\begin{abstract}
The surging development of Artificial Intelligence-Generated Content (AIGC) marks a transformative era of the content creation and production. Edge servers promise attractive benefits, e.g., reduced service delay and backhaul traffic load, for hosting AIGC services compared to cloud-based solutions. However, the scarcity of available resources on the edge pose significant challenges in deploying generative AI models. In this paper, by characterizing the resource and delay demands of typical generative AI models, we find that the consumption of storage and GPU memory, as well as the model switching delay represented by I/O delay during the preloading phase, are significant and vary across models. These multidimensional coupling factors render it difficult to make efficient edge model deployment decisions. Hence, we present a collaborative edge-cloud framework aiming to properly manage generative AI model deployment on the edge. Specifically, we formulate edge model deployment problem considering heterogeneous features of models as an optimization problem, and propose a model-level decision selection algorithm to solve it. It enables pooled resource sharing and optimizes the trade-off between resource consumption and delay in edge generative AI model deployment. Simulation results validate the efficacy of the proposed algorithm compared with baselines, demonstrating its potential to reduce overall costs by providing feature-aware model deployment decisions.
\end{abstract}
%Specifically, taking into account the heterogeneous deployment features of models in defining the objective function, a model-level deployment decision selection algorithm is proposed.

\section{Introduction}
Recently, the emergence of artificial intelligence-generated content (AIGC) has introduced a transformative paradigm in digital data creation and production. It is able to automatically generate contents that rival the quality of traditional contents, such as professionally-generated content and user-generated content \cite{xu2024unleashing,huang2024joint}. Consequently, generative AI models providing AIGC services have underscored a compelling need for efficient processing of end-user requests.

The primary infrastructure for AIGC services are cloud servers, providing users access to generative AI models through the network. Nevertheless, the drawback lies in the high delay and backhaul traffic load associated with the remote nature of cloud-based services \cite{zhang2018caching}. 
A noteworthy alternative arises from the utilization of edge servers equipped with computing facilities like graphics processing units (GPUs), thereby providing further feasibility and scalability for mobile AIGC networks. Essentially, the physical proximity between the user and the service provider enables users to access AIGC services with ultra-low delay, enhanced privacy protection, reduced bandwidth consumption, improved energy efficiency, etc., compared to cloud-based solutions \cite{yang2023adaptive,ogden2021many, kong2023edge}. 
Many existing research works advocate storing the features of inputs and task results at the edge server for possible reuse for future requests, thus facilitating low-delay services \cite{drolia2017cachier, yang2017dynamic}. 
However, storing results may not be effective to meet the demands of edge AIGC service, as it is difficult to satisfy customized interaction requirements from different users.
The request preference for AIGC services exhibits variability, wherein edge servers in diverse locations experience fluctuations in service types. For instance, edge nodes located within universities may encounter more demands for text generation services, while those close to business corporates might experience an upsurge in requests related to text to image services.
Therefore, elastic deployment of models at the edge server emerges as a promising approach, enabling the provision of real-time AIGC services without reliance on cloud servers. This approach dynamically adjusts deployment decisions to satisfy diverse user requests, while taking full advantage of edge servers.

Deploying models inevitably faces challenges associated with limited storage and GPU memory availability on the edge. Unlike cloud servers having ample resources to accommodate all generative AI models for providing AIGC services, the constrained resources of edge servers render it impractical to deploy all models simultaneously. Thereby, it gives rise to the issue of model miss: the model required to response to current user request is not deployed at the edge server. 
Subsequently, the edge server is compelled to download model from the cloud server and preload it into GPU memory, incurring additional delay and resource costs akin to those observed in content delivery networks (CDNs). Unlike the homogeneous attributes of pages or contents in CDNs, generative AI models exhibit heterogeneity in terms of resource consumption and service delay, including factors such as storage and GPU memory requirements, transmission delay, preloading delay and inference delay, with input/output (I/O) delay being a particularly overlooked factor in existing research \cite{ogden2021many,xu2023joint}. Thus, the selection of models to be deployed and the quantification of costs for model-level deployment become non-trivial.

Generative AI models are typically large in size, necessitating careful preloading and construction for their parameters and neural network structures. Stemming from these features, an efficient deployment solution tailored for generative AI models is not yet available. Consequently, common management approaches falter in handling these models, leading to unpredicted delay and suboptimal resource efficiency. With the proliferation of generative AI models, the complexity of managing them escalates, posing a significant challenge in addressing a multitude of model requests on a shared infrastructure. In particular, there is an urgent need for an edge deployment decision selection solution that can comprehensively consider the limited resources of edge server, as well as the heterogeneous demands of models and requests. 

To address aforementioned challenges, in this paper, we investigate the deployment of generative AI models at the edge server to empower mobile AIGC networks. We present a collaborative edge-cloud deployment framework in which multiple generative AI models can efficiently utilize shared resources in a cost-effective manner. In addition, to achieve the trade-off between resource and delay, we design a model-level deployment decision selection algorithm to manage generative AI models, which fully takes into account the heterogeneous characteristics including resource consumption and service delay of each model, as well as the impact of the request arrival rate on the deployment decisions. This ensures that the resource-constrained edge server is capable of providing real-time AIGC services at the lowest cost. The main contributions of this paper are summarized as follows.
%%%%%%%%%%%%%
\begin{itemize}
    \item We explore heterogeneous features of generative AI models, e.g., resource consumption and service delay. An analysis of the I/O delay intrinsic to models is presented.
    \item We formulate an edge model deployment problem as an optimization problem. Based on the heterogeneity of model features and the request arrival of AIGC services, we design a feature-aware model-level deployment decision selection algorithm to solve the problem under a collaborative edge-cloud deployment framework.
    \item Our proposed algorithm achieves a better trade-off between resource and delay compared to other baselines under various system settings, in which the algorithm not only reduces the average cost, but also maintains robustness under dynamic request arrival rates.
\end{itemize}
%Simulation experiments demonstrate that the proposed algorithm can reduce the total system cost by leveraging the distinctive characteristics inherent in generative AI models.

The remainder of this paper is organized as follows. We present our motivation in Section II. Section III explains our system model and problem formulation, as well as our proposed algorithm. In section IV, we demonstrate the evaluation results.Finally, we conclude our paper in Section V.
\iffalse
\begin{figure}[t]
    \centering
    \includegraphics[width=1\linewidth]{figure/overview.png}
    \caption{An overview of Edge-Cloud Collaboration Based AIGC Services.}
    \label{overview}
    %\vspace{-0.5cm}
\end{figure}
\fi
\section{Motivation}
In this section, to minimize the impact of the demand features of generative AI models on resource and delay, we are motivated to explore these features in depth.
\subsection{Resource Consumption of Generative AI Models}
\begin{figure}[t]
    \centering
    \includegraphics[width=\linewidth]{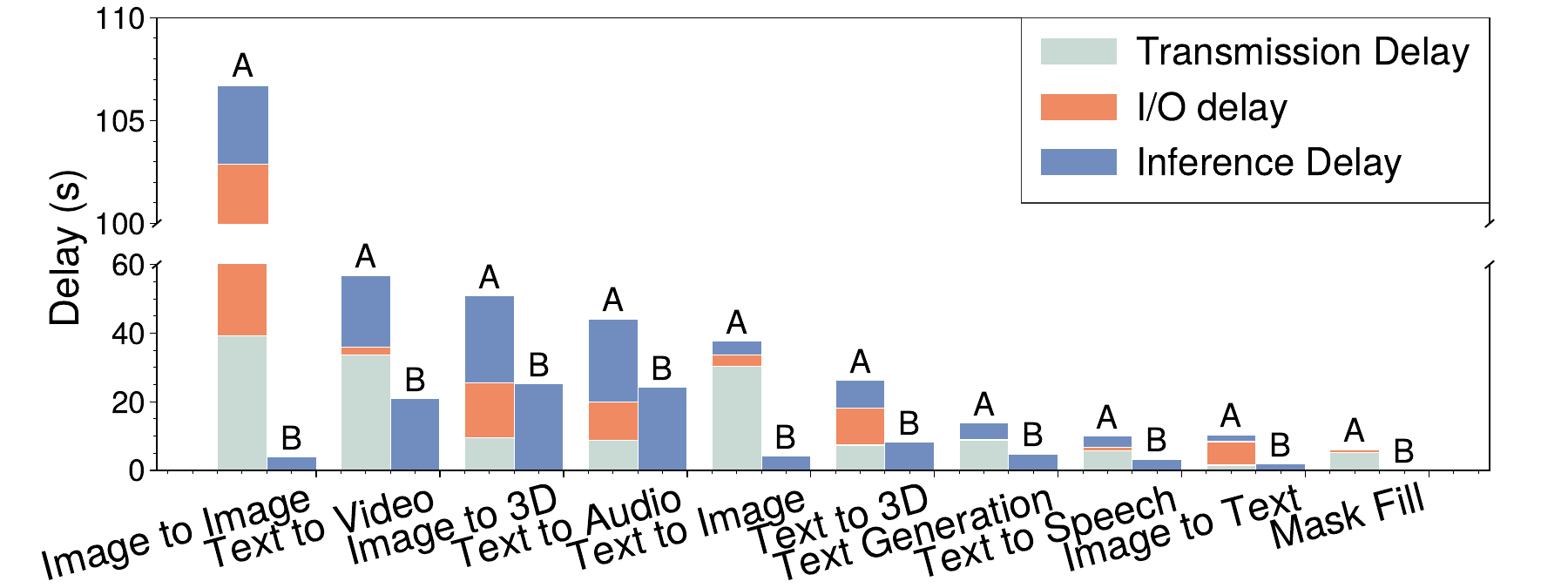}
    \vspace{-0.6cm}
    \caption{Service delay of undeployed model A and deployed model B within a data rate of 10 Gbps between the cloud and the edge.}
    \label{fig:motivation2}
    \vspace{-0.4cm}
\end{figure}

Generative AI models are esteemed for the ability of customization and high-quality generated contents. This efficacy stems from their extensive parameters and subtle neural network structures \cite{xiangxiang2024axiomvision}. Consequently, these models entail considerable sizes, mandating ample deployment storage space.

Moreover, generative AI models should be preloaded into GPU memory for subsequent inference, inevitably occupying resources, particularly the scarce GPU memory.
To characterize the resource demands of models, we conduct a motivating observation covering ten typical AIGC services, such as Text to Image, Text Generation, Text to Music, etc. Each of those services leverages a corresponding lightweight generative AI models, including Stable Diffusion V1-4, GPT2, MusicGen \cite{Rombach_2022_CVPR,radford2019language,copet2024simple,jun2023shap,luo2023videofusion,xu2022image,pratap2023mms,meng2021sdedit,DBLP:journals/corr/abs-1810-04805}. The storage and GPU memory demands of these models vary significantly, ranging from 2 GB to 50 GB for the former and from 1 GB to 6 GB for the latter. Consequently, the storage and GPU memory requirements of generative AI models show significant heterogeneity.

\subsection{I/O Delay of Generative AI Models}                    
To obtain generative content, the phases of model transmission, preloading, and inference contribute to service delay. Notably, the preloading phase incurs pronounced delay overhead. During this phase, model parameters, execution structures, and relevant data are fetched from the storage disks of the CPU to the GPU memory through I/O interfaces, e.g., the PCIe interface, akin to the cold-start delay in Function-as-a-Service. However, this aspect has been scarcely considered in existing studies on the deployment of generative AI models.

We explore the service delay of each model at the edge server over 30 load-offload cycles with 10 executions per cycle within a data rate of 10 Gbps between the cloud and the edge. As shown in Fig. \ref{fig:motivation2}, the first service delay is much longer than that of the preloaded ones. This is because for an undeployed model, its service delay consists of transmission delay, I/O delay, and inference delay, while the subsequent one consists of only the inference delay. Besides, it can also be found that the transmission delay and I/O delay of the model have a significant impact on the overall service delay. Therefore, when making decision of deploying generative AI models at the edge server, it is essential to consider not only the resource consumption, but also the first service delay. 

Motivated by these observations, it is crucial to make proper decisions based on model-specific features, including storage and GPU memory consumption, as well as deployment delays, particularly those caused by I/O read operations.

\begin{figure}[]
    \centering
    \includegraphics[width=1\linewidth]{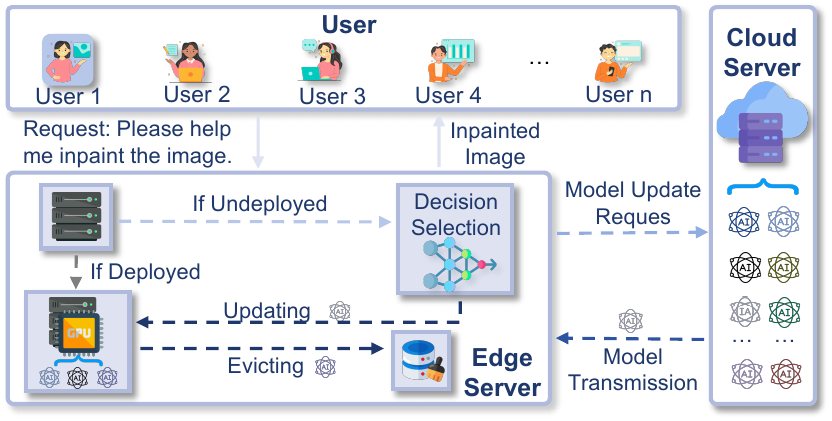}
    \vspace{-0.6cm}
    \caption{An overview of the collaborative edge-cloud deployment framework.}
    \label{fig:workflow}
    \vspace{-0.4cm}
\end{figure}

\section{System Model and Problem Formulation}
In this section, we present a collaborative edge-cloud deployment framework tailored for generative AI models, followed by decision model and cost model. Finally, we formulate an optimization problem and design a model-level deployment selection algorithm to solve the problem.
\subsection{System Model}
%Consider a collaborative edge-cloud deployment framework for generative AI models in Fig. \ref{fig:workflow}. The framework comprises a cloud server hosting all generative AI models, along with a number of edge servers tasked with deploying and updating a subset of models, catering to AIGC services. 
Consider a collaborative edge-cloud deployment framework in Fig. \ref{fig:workflow}. The framework comprises a cloud server hosting generative AI models, edge servers responsible for deploying and updating a subset of models, and users requesting services. Upon receiving a user request at the edge server, if the required model is already deployed, the service is promptly delivered. Otherwise, the system undergoes a model deployment decision-selecting process. Following this, the selected model is deployed and updated at the edge, after which the service is responded.
For the cloud server, we presume a library of generative AI models, denoted by $ \mathcal{M}= \{1,2,...,M\}$, are available. For each model $m\in \mathcal{M}$, it is characterized by its size, $s_m$, GPU memory consumption, $g_m$, energy consumption, $e_m$, I/O delay, $d_m$, and inference time, $i_m$.
 
\subsection{Decision Model}
To achieve rapid service response and minimize resource consumption, it is essential for each edge server to efficiently deploy and update models. Consequently, selecting decision regarding the deployment of models holds paramount importance. Without loss of generality, we partition a period of request time into $T$ time slots of equal length $\tau$, denoted sequentially as $\mathcal{T} = \{1, 2, 3, ..., T\}$. Let $a_{m,t} \in \{0, 1\}$ represent a binary decision variable, and $a_{m,t} = 1$  means that model $m$ is deployed at the edge server at time slot $t$. The deployment decision is indicated as a vector $\mathbf{A}_t = [a_{1,t}, a_{2,t}, \dots, a_{M,t}]$.

A certain AIGC service can be rapidly provisioned at the edge server if required model is available to be deployed within the constraints of resources, including storage capacity, GPU memory, and energy. Let $C$ denote the storage capacity of each edge server, and it is not sufficient to accommodate all generative AI models. Consequently, the deployment decision is bound by the constraint:

\begin{equation}\label{1}
    \sum_{m\in \mathcal{M}} a_{m,t}s_m \le C, \quad \forall t \in \mathcal{T}.
\end{equation}

%   Equation (\ref{1}) ensures that the total storage demand does not exceed the capacity limit $C$ across all time slots $t$.

Moreover, according to the observations in Section II, the deployment decision of generative AI models is further constrained by the GPU memory at the edge server, as models with neural network structures and parameters need to be preloaded into the GPUs for prompt response to AIGC services. Let $G$ be the available GPU memory to preload models at the edge server, it holds that:

\begin{equation}\label{2}
    \sum_{m\in \mathcal{M} } a_{m,t}g_m \le G, \quad \forall t \in \mathcal{T}.
\end{equation}

%Here, it constraint ensures that the total GPU memory demand does not exceed the allocated capacity $G$ across all time slots $t$.

Given that preloading the model into the GPU incurs the energy consumption of the edge server, the energy consumption is subjected to the overall energy constraint, which can be expressed as:
\begin{equation}\label{3}
   \gamma + \sum_{m\in \mathcal{M}}  e_{m}a_{m,t} \le E, \quad \forall t \in \mathcal{T},
\end{equation}
where $E$ denotes the energy budget controlled by the rated power of each edge server, with $\gamma$ representing the static power consumption irrespective of workload. 
\subsection{Cost Model}
The edge server caters for AIGC services by making user requested models available, which highly relies on model deployment decisions. When the model miss event occurs, the deployment decision is selected to update models based on both the switching cost and the resource cost.

\subsubsection{Switching Cost}
We define switching cost as the delay of deploying new models due to the model miss event. Switching models at $t$-th time slot may cause a longer response time in the future, because if a deployed model that is currently selected not to be deployed is requested again, it will result in the transmission delay, preloading I/O delay, and inference delay in the future, which are components of switching cost.

%Upon receipt of each incoming AIGC request that necessitates an undeployed model, the edge server triggers a model update request to the cloud server, entailing a update request delay denoted $L_r$. 

Firstly, the model undergoes transmission from the cloud server to the edge server, incurring a transmission delay:
\begin{equation}\label{eq:transmission}
    L_{1}(\mathbf{A}_t)=\sum_{m\in \mathcal{M}}\frac{s_m}{B }\mathbf{I}(a_{m,t} < a_{m,t-1}),
\end{equation}
where $B$ denotes the bandwidth allocated for model transmission.
$\mathbf{I}(\cdot)$ is the indicator function. $\mathbf{I}(a_{m,t} < a_{m,t-1})$ means the deployed model $m$ is decided to be evicted at present.

Then, to ensure prompt response for AIGC services, the model is preloaded into GPU memory through the I/O interface of the edge server, facilitating subsequent inference operations. The preloading I/O delay can be given by:
\begin{equation}\label{eq:preloading}
    L_{2}(\mathbf{A}_t)=\sum_{m\in \mathcal{M}}  d_m\mathbf{I}(a_{m,t} < a_{m,t-1}).
\end{equation}

Finally, $i_m$ represents the inference time of generative AI model $m$ at the edge server. The inference time can be regarded as a constant under the condition that the inputs and the runtime resource allocation is deterministic. This phase incurs a inference delay, which is computed as
\begin{equation}\label{eq:inference}
    L_{3}(\mathbf{A}_t)=\sum_{m\in \mathcal{M}}  i_m\mathbf{I}(a_{m,t} < a_{m,t-1}).
\end{equation}

The dynamic arrival of service requests can also influence the switching cost. While a model $m$ has longer transmission delay, I/O delay or inference delay, if it remains unrequested for an extended period, deploying it at the edge may increase the instances of missing a model with shorter delays but higher access frequency, resulting in longer average delay. Hence, the user's dynamic request arrival rate is crucial to consider. Thus, we define the active cycle $\beta_m$ for each model $m$, as the number of requests until model $m$ is subsequently requested. A large $\beta_m$ means that model may be unlikely to be requested again in the future. Therefore, the total switching cost of the edge server is formulated as:
\begin{equation}\label{eq:switching}
    L_{t}(\mathbf{A}_t)= \frac{L_{1}(\mathbf{A}_t)+L_{2}(\mathbf{A}_t)+L_{3}(\mathbf{A}_t)}{\sum_{m \in \mathcal{M}}\beta_m\mathbf{I}(a_{m,t} < a_{m,t-1})},
\end{equation}
which aims to simultaneously reduce the number of model miss events by utilizing $\beta_m$ and prioritize the deployment of models with higher delays based on the value of $L_{1}(\mathbf{A}_t)+L_{2}(\mathbf{A}_t)+L_{3}(\mathbf{A}_t)$.
Given the negligible nature of the model update request delay compared to other delay components, it can be disregarded in our design.

%where $L_r$ is a update request delay. Given the negligible nature of it relative to other delay components, typically on the order of a few tens of milliseconds, it is regarded as constant and thus disregarded in our design.

\subsubsection{Resource Cost}
The concept of resource cost pertains to the utilization of resources of each edge server. Each deployment decision influences the allocation of resources consumed by models deployed at time slot $t$. According to the observations in Section II regarding the resource demands of generative AI models, the resource cost of each decision consists of storage capacity cost and GPU memory costs.

When deploying a generative AI model at the edge server, it inherently consumes storage space for storing a considerable number of model parameters and related data. The storage cost can be expressed as:
\begin{equation}\label{eq:storage}
    R_{1}(\mathbf{A}_t)=\sum_{m\in \mathcal{M}}  a_{m,t}s_m.
\end{equation}

Once the model is downloaded and stored, model parameters are read from hard disk to GPU memory and the execution graph is configured during the preloading phase, thus occupying a certain amount of GPU memory for executing subsequent inference as soon as the service request arrives. Hence, GPU memory consumption cost can be quantified as:
\begin{equation}\label{eq:gpu}
    R_{2}(\mathbf{A}_t)=\sum_{m\in \mathcal{M}}  a_{m,t}g_m.
\end{equation}

Consequently, the total resource cost comprising both storage capacity and GPU memory cost is formulated as:
\begin{equation}\label{eq:resource}
    R_{t}(\mathbf{A}_t)= R_{1}(\mathbf{A}_t)+wR_{2}(\mathbf{A}_t),
\end{equation}
where $w$ is a weight parameter accounting for the balance between the storage cost and the GPU memory cost.
\begin{algorithm}[t]
    \caption{GA-based Deployment Decision Selection}
    \label{alg:1}
    \hspace*{0.02in}{\bf Input:} $C, G, E, B,\tau, \gamma, s_m, g_m, e_m, d_m, i_m, \beta_m$ ($\forall m \in \mathcal{M}$) \\
    \hspace*{0.02in}{\bf Output:} Deployment decision $\mathbf{A}_t =  [a_{1,t}, a_{2,t}, \dots, a_{M,t}]$
    \begin{algorithmic}[1]
        \State Set probabilities $p_1$, $p_2$ and generation counter $n = 1$
        \For {$k=1$ to $K$}
            \State Randomly generate $\mathbf{A}^{(k,1)}_{t} = [a^{(k,1)}_{1,t}, a^{(k,1)}_{2,t}\dots a^{(k,1)}_{M,t}]$ 
            \State Calculate $C^{(k,1)}_{t} = \mu_LL_{t}(\mathbf{A}^{(k,1)}_{t}) + \mu_R R_{t}(\mathbf{A}^{(k,1)}_{t})$
            \State Calculate $F^{(k,1)}_{t} = {(\frac{C_{t,max}^{(k,1)}-C_{t}^{(k,1)}}{C_{t,max}^{(k,1)}-C_{t,min}^{(k,1)}})}^2$
        \EndFor
        \While{$n \le N $}
            \State Normalize fitness: $P^{(k,n)} = \frac{F^{(k,n)}}{\sum_{k = 1}^{K} F^{(i,n)}}$
            \State Select parents: $\mathbf{A}_{t}^{(k,n+1)} \leftarrow RandomChoose(\mathbf{A}_t^{(k,n)})$
            \For {$k=1$ to $K$}
                \State Crossover $p_1$: $\mathbf{A}_t^{(k,n)} \leftarrow C(\mathbf{A}_t^{(k^{\prime},n)}, {\mathbf{A}_t^{(k^{\prime \prime},n)}})$
                \State Mutation $p_2$: $\mathbf{A}_t^{(k,n)} \leftarrow M(\mathbf{A}_t^{(k,n)})$
                \State Calculate $C_t^{(k,n)} = \mu_LL_{t}(\mathbf{A}_t^{(k,n)}) + \mu_R R_{t}(\mathbf{A}_t^{(k,n)})$
                \State  Calculate $F^{(k,n)}_{t} = {(\frac{C_{t,max}^{(k,n)}-C_{t}^{(k,n)}}{C_{t,max}^{(k,n)}-C_{t,min}^{(k,n)}})}^2$
            \EndFor
            \State Update generation counter: $n = n+1$
        \EndWhile
        \State Assign $\mathbf{A}_t$ as argmax($F(\mathbf{A}_t^{(k,n)})$)
    \end{algorithmic}  
\end{algorithm} 

\subsection{Problem Formulation}
To reduce service delay and edge resource consumption, we formulate the problem as an optimization problem. Therefore, we concurrently address both model switching cost and resource cost at once over a period of requesting time $T$. The problem can be formulated as follows:
\begin{subequations}
\begin{equation}\label{eq:p}
   \mathcal{P}: \min_{\mathbf{A}_t} \frac{1}{T}\sum_{t \in \mathcal{T}}[\mu_LL_{t}(\mathbf{A}_t ) + \mu_R R_{t}(\mathbf{A}_t)]
\end{equation}
\begin{equation}\label{eq:c1}
\text{s.t.}\quad (1), (2), (3), \qquad \qquad
\end{equation}
\begin{equation}\label{eq:c2}
\begin{aligned}
a_{m,t} \in \{0, 1\}, \forall m \in \mathcal{M}, t \in \mathcal{T},
\end{aligned}
\end{equation}
\end{subequations}
where $\mu_L$, $\mu_R$ denote the non-negative weight parameters of service delay and resource consumption, which can be tailored based on the preference settings of the edge server or according to the multiple criteria decision making theory \cite{wallenius2008multiple}. Eq. (\ref{eq:c1}) delineates the resource constraints pertinent to the edge server, encompassing storage and GPU memory, as well as energy capacity. Eq. (\ref{eq:c2}) specifies the binary nature of the decision variables governing model updates. Problem $\mathcal{P}$ exhibits exponential time complexity, rendering brute force approaches impractical, particularly as the number of generative AI models escalates. Hence, a lightweight algorithm should be designed to solve problem $\mathcal{P}$.

\iffalse
\begin{figure*}[t]
    \centering
    \subfigure[]{
    \vspace{-0.8cm}
    \label{fig:storage}
    \begin{minipage}[t]{0.5\linewidth}
    \centering
    \includegraphics[width=1 \linewidth]{figure/cache.pdf}
    \end{minipage}%
    }%
    \subfigure[]{
    \vspace{-0.8cm}
    \label{fig:gpu}
    \begin{minipage}[t]{0.5\linewidth}
    \centering
    \includegraphics[width=1 \linewidth]{figure/gpu.pdf}
    \end{minipage}
    }%
    \centering
    \vspace{-0.3cm}
    \caption{Average cost versus the number of (a) storage capabilities and (b) GPU memories under the same arrival rate and the number of the time slot.}
    \label{fig:result1}
    \vspace{-0.5cm}
\end{figure*}
\fi

\begin{figure*}[htbp]
\centering
\subfigure[]
{
    \begin{minipage}[b]{0.32\linewidth}\label{fig:cache1}
        \centering
        \includegraphics[width=1 \linewidth]{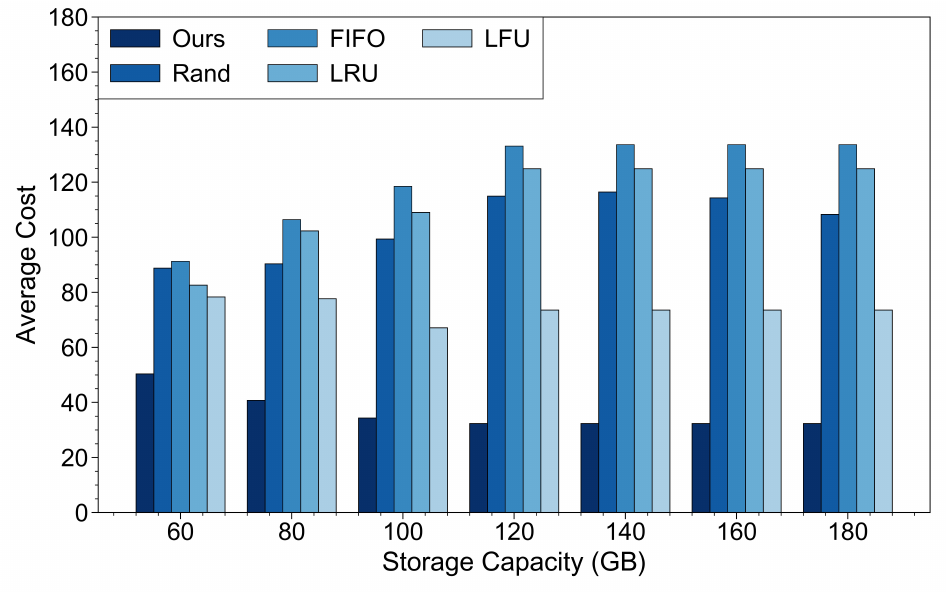}
    \end{minipage}
}
\hspace{-0.2in}
\subfigure[]
{
    \begin{minipage}[b]{0.17\linewidth}\label{fig:cache2}
        \centering
        \includegraphics[width=1 \linewidth]{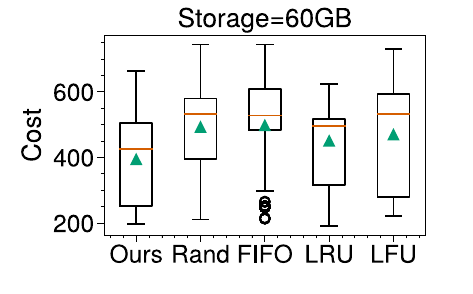} \\
        \includegraphics[width=1 \linewidth]{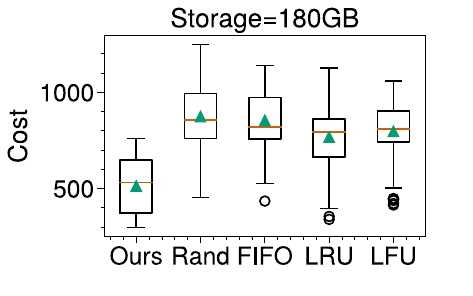}
    \end{minipage}
}
\hspace{-0.2in}
\subfigure[]
{
    \begin{minipage}[b]{0.32\linewidth}\label{fig:gpu1}
        \centering
        \includegraphics[width=1 \linewidth]{figure/cache1.pdf} 
    \end{minipage}
}
\hspace{-0.2in}
\subfigure[]
{
    \begin{minipage}[b]{0.17\linewidth}\label{fig:gpu2}
        \centering
        \includegraphics[width=1 \linewidth]{figure/cache2.pdf} \\
        \includegraphics[width=1 \linewidth]{figure/cache3.pdf}
    \end{minipage}
}
\centering
\vspace{-0.3cm}
\caption{Average cost and cost statistics versus the number of (a)(b) storage capabilities and (c)(d) GPU memories under the same arrival rate and the number of the time slot.}
\label{fig:result1}
\vspace{-0.4cm}
\end{figure*}

\subsection{Model-Level Deployment Decision Selection Algorithm}
To address problem $\mathcal{P}$, we propose a genetic algorithm (GA)-based approach to select effective model-level deployment decisions. By leveraging GA's inherent traits of convergence and efficiency, we achieve a reduction in computational complexity while maintaining robust convergence, with the goal of optimizing the objective fitness function:
\begin{equation}\label{eq:f}
        F = {(\frac{C_{max}-C}{C_{max}-C_{min}})}^2,
\end{equation}
where $C$ is the decision cost according to Eq. (\ref{eq:p}). The algorithm is sketched in Algorithm \ref{alg:1}. 
Each individual represents a deployment decision. 
In lines 2-5, the first generation in the population is formed by randomly generating $K$ individuals. Lines 6-14 orchestrate an iterative update of the population. The $K$ individuals are selected based on their fitness evaluated by Eq. (\ref{eq:f}).
Crossover occurs between two individuals of the same generation with a probability $p_1$, implying that crossover offspring inherits some of the genes of the parent, combining the deployment decisions of the models from parents. Additionally, each individual may undergoing mutation with a probability $p_2$. If a gene mutates from 0 to 1, it signifies that the model decision represented by the gene will change from eviction to deployment.
The while loop continues until the generation counter reaches a pre-established threshold denoted by $N$, or the result convergences.

\section{Performance Evaluation}
In this section, we evaluate the performance of the proposed algorithm. Firstly, we provide the experimental settings. Secondly, we evaluate the effectiveness of the proposed algorithm with different resources on the edge. Finally, the robustness of our algorithm is presented.
\subsection{Experimental Settings}
\subsubsection{Parameter Settings}
We consider a collaborative edge-cloud deployment system with 100 time slots in our experiments. The request arrival of AIGC services per time slot follows a Poisson process with a rate of $\lambda_m$. The request history of each model is able to be used to estimate the arrival time of the next request \cite{ogden2021many,wu2024Characterizing}. Specifically, for each model $m$, $\beta_m$ can be approximate by $\frac{1}{\lambda_m}$. We consider 10 types of AIGC services and the corresponding representative generative AI models illustrated in Section II. Other important parameters of the experiments are listed in Table \ref{tab:test2}.

\subsubsection{Performance Baselines}
We compare the model-level decision selection algorithm tailored for generative AI models with following baselines.
\begin{itemize}
\item Random (Rand): It randomly selects models to be evicted.
\item First-in-first-out (FIFO): It evicts the model with the longest deployment time.
\item Least recently used (LRU): It evicts the least recently accessed deployed model, based on the recently accessed models are more likely to be requested again.
\item Least frequently used (LFU): It evicts models with the least frequent requests, considering that the most frequently accessed models are more likely to be requested.
\end{itemize}
\begin{table}[t]
	\begin{center}
		\caption{Parameters Setting} \label{tab:test2}
		\vspace{-0.05in}
		\scalebox{1}{
			\begin{tabular}{c|c}
				%\begin{tabular}{m{0.21\linewidth} | b{0.72\linewidth}}
				\hline 
				\textbf{Parameter} & \textbf{Value}  \\
				\hline 
				Storage space requirement of the model $m$, $s_m$    & [2,50] GB \\
				\hline
				GPU memory requirement of the model $m$, $g_m$          & [1,6] GB\\
				\hline
				Energy consumption of the model $m$, $e_m$    & [0.0025,0.5] kW\\
				\hline
			I/O delay of the model $m$, $d_m$      & [0.3,60] s\\
                    \hline
                    Inference delay of the model $m$, $i_m$          &[0.05,30] s \\
				\hline
				Storage capacity of the edge server, $C$        & [60,180] GB   \\
				\hline
				GPU Memory of the edge server, $G$      & [6,18] GB \\
                    \hline
				Energy capacity of the edge server, $E$       & 1 kW  \\
                    \hline
				Static power of the edge server, $\gamma$         & 0.3 kW  \\
				\hline
                    %Network bandwidth, $B$          & 10 Gbps \\
				%\hline
                    %Weight parameter, $w$          & 10 \\
				%\hline
                    %Weight parameter, $\mu$          & 2 s/GB \\
				%\hline
		\end{tabular}}
		\vspace{-0.37in}
	\end{center}
\end{table}
\subsection{Performance Comparison}
\iffalse
\begin{figure*}[t]
    \centering
    \subfigure[]{
    \label{fig:slot1}
    \begin{minipage}[t]{0.5\linewidth}
    \centering
    \includegraphics[width=1 \linewidth]{figure/slot1.pdf}
    \end{minipage}%
    }%
    \subfigure[]{
     \vspace{-0.3cm}
    \label{fig:slot2}
    \begin{minipage}[t]{0.5\linewidth}
    \centering
    \includegraphics[width=1 \linewidth]{figure/slot2.pdf} %%\lamda=1
    %\caption{fig2}
    \end{minipage}
    }%
    \centering
    \vspace{-0.3cm}
    \caption{Average cost versus the number of time slots under (a) different arrival rates and (b) the same arrival rate.}
    \label{fig:result2}
    \vspace{-0.5cm}
\end{figure*}
\fi

\begin{figure*}[htbp]
\centering
\subfigure[]
{
    \begin{minipage}[b]{0.32\linewidth}\label{fig:slot11}
        \centering
        \includegraphics[width=1 \linewidth]{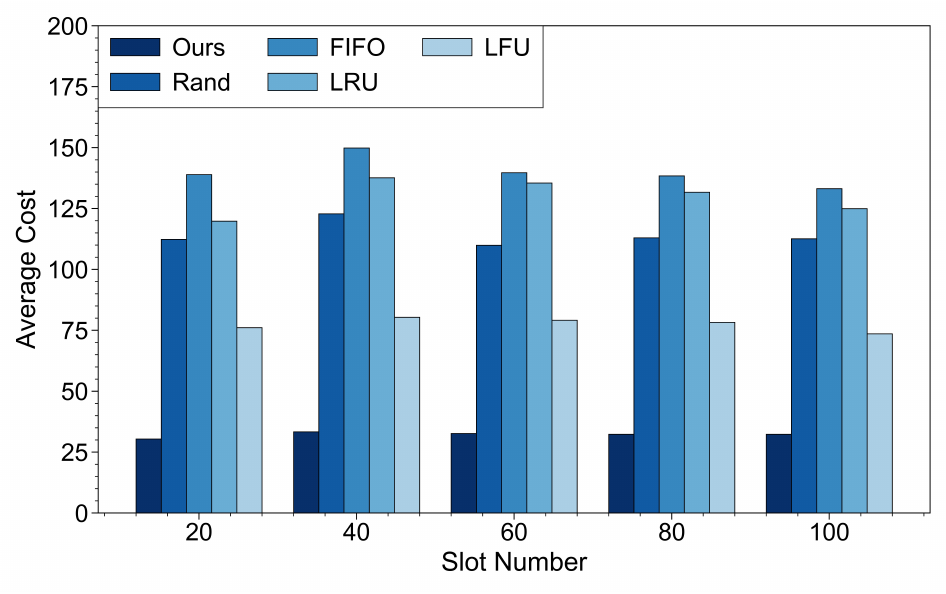}
    \end{minipage}
}
\hspace{-0.2in}
\subfigure[]
{
    \begin{minipage}[b]{0.17\linewidth}\label{fig:slot12}
        \centering
        \includegraphics[width=1 \linewidth]{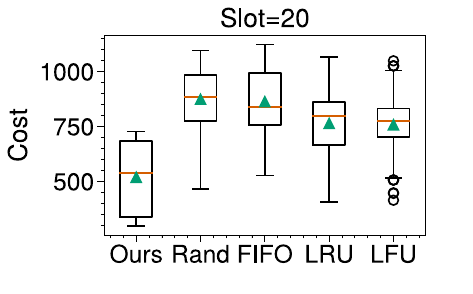} \\
        \includegraphics[width=1 \linewidth]{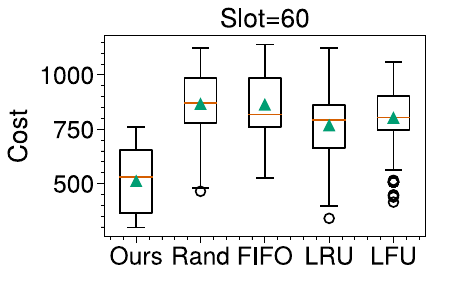}
    \end{minipage}
}
\hspace{-0.2in}
\subfigure[]
{
    \begin{minipage}[b]{0.32\linewidth}\label{fig:slot21}
        \centering
        \includegraphics[width=1 \linewidth]{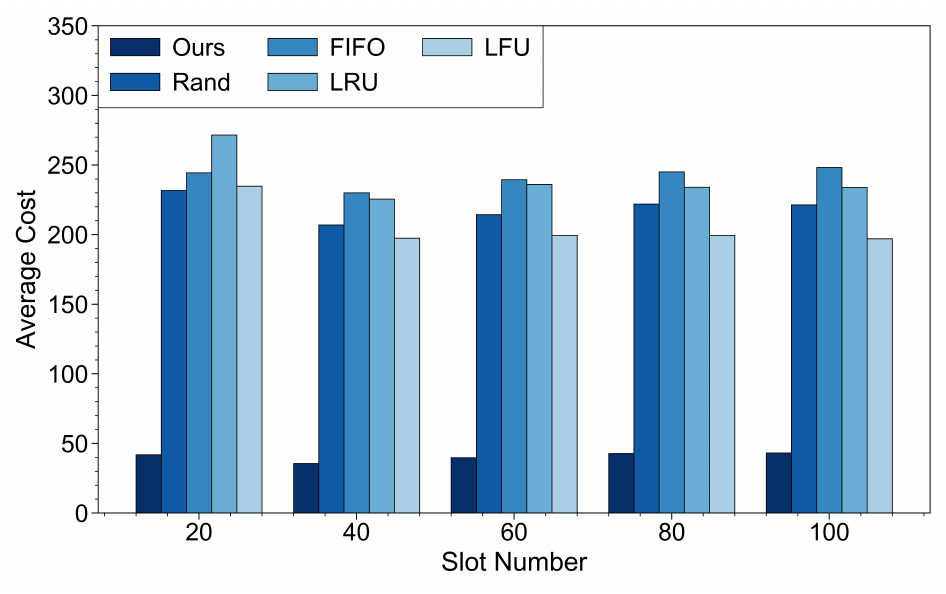} 
    \end{minipage}
}
\hspace{-0.2in}
\subfigure[]
{
    \begin{minipage}[b]{0.17\linewidth}\label{fig:slot22}
        \centering
        \includegraphics[width=1 \linewidth]{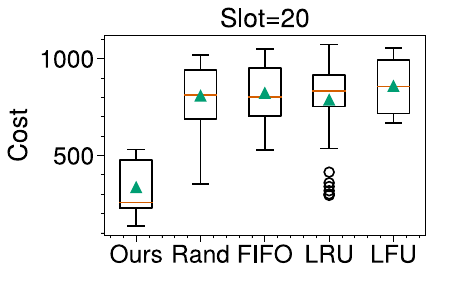} \\
        \includegraphics[width=1 \linewidth]{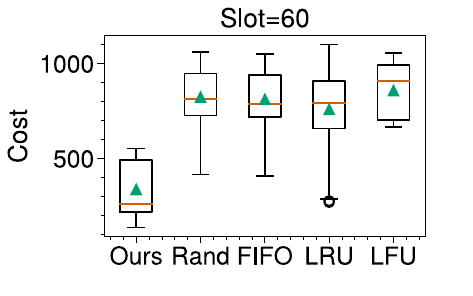}
    \end{minipage}
}
\centering
\vspace{-0.3cm}
\caption{Average cost versus the number of time slots with (a) (b) varying arrival rates based on popularities across models, and (c) (d) the same arrival rate.}
\label{fig:result2}
\vspace{-0.4cm}
\end{figure*}

The effectiveness of our proposed algorithm is evaluated through a comparative analysis of the average cost under varying resource constraints of the edge server, i.e., storage capacity and GPU memory. The results are shown in Fig. \ref{fig:result1}. Fig. \ref{fig:cache1} illustrates the average cost of each method as the storage increases while keeping the GPU memory constant at 12 GB. Similarly, Fig. \ref{fig:gpu1} depicts the trend of the average cost as the GPU memory changes with the storage capacity at 120 GB. Notably, the average costs of the proposed algorithm and LFU gradually decrease and ultimately converge to stable values, contrasting with the escalating cost trends observed in other baselines as the resource allocation increases. These observations validate the superiority of the proposed algorithm to facilitate efficient deployment of generative AI models within the limited resources at the edge server. 
More statistics of the cost  under different resources are also presented in Fig. \ref{fig:cache2} and Fig. \ref{fig:gpu2}, highlighting the advantage of the proposed algorithm. In particular, when storage is set at 180 GB or GPU memory at 18 GB, the cost of the proposed algorithm is lower than that of almost any other baselines.
Our algorithm always opts for the most cost-effective decision, thus maintaining the lowest average cost under across diverse resource constraints. Specifically, the proposed algorithm reduces the average cost from 56.06\% to 76.73\% compared to baselines when storage and GPU memory are set at 120 GB and 12 GB, respectively. Hence, the algorithm proposed in our paper strikes a delicate balance between resource consumption and service delay.

We then demonstrate the robustness of the proposed algorithm within dynamic service environments, considering varying numbers of time slots and request arrival rates determined by model popularities. As shown in Fig. \ref{fig:slot11} and Fig. \ref{fig:slot12}, when the number of time slots varies, we observe the trends and statistics of the average cost among different algorithms. Clearly, the average cost of our proposed algorithm consistently maintains a stable and minimal level. Moreover, we standardize the user request arrival rate, $\lambda$, across all models and generate the corresponding request arrival statistics. The results presented in Fig.  \ref{fig:slot21} and Fig. \ref{fig:slot22} show that our algorithm achieves the lowest average cost compared to baselines. On the one hand, the robust performance of our algorithm stems from its long-term perspective when confronts with prolonged AIGC requests at the edge server. Our algorithm strategically weighs the trade-off of deployment decisions, considering both current resource consumption and future delay at each time slot, thus ensuring optimal performance in the long run. On the other hand, most baselines deploy models based on simplistic patterns, such as the order or frequency of service requests, without adequately considering the inherent features of generative AI models observed in Section II. Consequently, these baselines exhibit subpar performance in scenarios where such simplistic patterns are not readily represent requirements.

\section{Conclusion}
In this paper, we have investigated the efficient deployment of generative AI models characterized by diverse demand levels of resource consumption and service delay on the edge. We have posited that with the burgeoning popularity of AIGC services, judicious deployment of models on the edge holds significant application potential. We have presented a collaborative edge-cloud deployment framework tailored for generative AI models, in which a novel feature-aware model-level deployment decision selection algorithm has been proposed to aim at minimizing service delay while adhering to the constraints of edge resource consumption by adapting to the specific features of generative AI models. 
%Extensive simulation experiments have shown that the proposed algorithm achieved the lowest average cost compared to baselines and maintained robustness under dynamic AIGC service requests. 
The proposed framework can be applied to mobile AIGC networks to enable users to access AIGC services with ultra-low service delay, as well as enhance well-organized resource consumption ability of the edge servers providing services.
For the future work, we will explore the joint edge-cloud workload scheduling and updating of AIGC services and models.
\section*{Acknowledgement}
The work was supported in part by the Natural Science Foundation of China under Grant 62001180, in part by the Young Elite Scientists Sponsorship Program by CAST under Grant 2022QNRC001, and in part by Hubei Provincial Natural Science Foundation of China under Grant 2024AFD413.

\footnotesize
\bibliographystyle{IEEEtran}
\bibliography{ref}

\end{document}